\documentclass[a4paper,twocolumn,twoside]{article}%
\usepackage{sectsty}

\sectionfont{\large}
\subsectionfont{\normalsize}

\usepackage[sort&compress,square,numbers]{natbib}
\newcommand{\bibfont}{\footnotesize}%
\usepackage{fancyhdr}
\renewcommand{\headrulewidth}{0.7pt}
\renewcommand{\footrulewidth}{0.7pt}
 \fancypagestyle{plain}{
    \renewcommand{\headrulewidth}{0pt}
    \renewcommand{\footrulewidth}{0pt}
    \fancyhf{}
}

\usepackage{subfigure}
\usepackage{latexcad}

\usepackage{graphicx}
\usepackage{amssymb}

\newtheorem{definition}{Definition}

\begin{document}
\title{\vspace{-9mm}\Large\bf Fuzzy Color Model and Clustering Algorithm for Color Clustering Problem\vspace{4mm}}
\author{{\normalsize Dae-Won Kim, \quad Kwang H. Lee \vspace{1.5mm}}\\
{\small Department of Computer Science, Korea Advanced Institute
of Science and Technology}\\
{\small Kusung-dong, Yusung-gu, 305-701, Daejon, Korea
\vspace{-4mm}}}
\date{} \maketitle
\renewcommand\thefootnote{\fnsymbol{footnote}}
\footnotetext{\footnotesize \vspace{-0.2cm}\\Proceedings of the
First International Conference on Information and Management
Sciences, Xi'An, China, May 27-31, 2002, pp.~\pageref{first}
-~\pageref{last}.\\ This work was supported by the Korea Science
and Engineering Foundation (KOSEF) through the Advanced
Information Technology Research Center(AITrc).}
\thispagestyle{empty} \label{first}

\noindent\textbf{Abstract:} The research interest of this paper is
focused on the efficient clustering task for an arbitrary color
data. In order to tackle this problem, we have tried to model the
inherent uncertainty and vagueness of color data using fuzzy color
model. By taking fuzzy approach to color modeling, we could make a
soft decision for the vague regions between neighboring colors.
The proposed fuzzy color model defined a three dimensional fuzzy
color ball and color membership computation method with two
inter-color distances. With the fuzzy color model, we developed a
new fuzzy clustering algorithm for an efficient partition of color
data. Each fuzzy cluster set has a cluster prototype which is
represented by fuzzy color centroid.


\vspace{-1mm}\section*{I. Introduction} \label{sec:introduction}
Color is the one of the most important features in our lives. It
is not easy task to effectively describe color in our language
even though color can be considered as a simple and intuitive
object~\cite{Lon94}\cite{Luk96}\cite{Zel98}. The research interest
of this paper is focused on the color clustering problem. For a
given set of color data and the number of clusters, the objective
is to partition the color set into homogenous color sub-partition.
This kind of color clustering task can be widely used for a
variety of applications, for example, color image segmentation.
The difficulties of this research include not only the lack of
correct color model which can describe the uncertain
characteristics of color, but also the lack of efficient
clustering algorithm which can deal with the vague color data.

All the literatures related to color modeling including RGB, HSV,
CMY, CIELAB color spaces have only handled the crisp
representation of color data. However, color has inherently
uncertain and vague characteristics. For the overlapping area
between two major colors, the absolute color classification is not
possible because it mainly depends on both the visual decision of
the observer and the surrounding color effects. Thus we have tried
to develop a new color model that can represent the color
uncertainty and vagueness, which results in a proposed fuzzy color
model. We modeled a color space with fuzzy-set theory and created
a notion of three dimensional fuzzy color ball. And we proposed
two color distance measures including the distance between a color
element and a fuzzy color and the distance between fuzzy colors.
With the distance measures, the computation method for color
membership value was defined.

In order to effectively partition the color data set, fuzzy
cluster analysis technique is best choice due to the ability of
dealing with color uncertainty. But most of previous fuzzy
clustering algorithms were designed for crisp pattern data, thus
we need to develop an new algorithm that can tackle the fuzzy data
clustering. We applied the proposed fuzzy color model to devise a
new approach in fuzzy clustering algorithm. Color is a qualitative
feature and we have to make a soft decision in color clustering
activity. Thus each color cluster set is prototyped by color
centroid value that is represented by fuzzy color ball. By
minimizing the pre-defined evaluation criteria, we could obtain
the near-optimal convergence in an iterative color clustering.

The remainder of this paper is organized as follows. Section II
gives the description of background knowledge about color and
previous works in cluster analysis. We describe our new fuzzy
color model and its color membership computation method in section
III , and the proposed fuzzy clustering algorithm with fuzzy color
model is covered in section IV. Finally we conclude our remarks in
section V.


\vspace{-1mm}\section*{II. Literature Review}
\label{sec:literature-review}

\subsection*{2.1\quad Fundamental Color Spaces}
\label{sec:fundamental-color-spaces}

In this section, we'll cover the three fundamental color spaces,
including RGB, CMY, Munsell, and CIE uniform color systems. We
will briefly describe the basic concept and its color models.

\subsubsection*{2.1.1\quad RGB and CMY Color Spaces}
\label{sec:rgb-and-cmy-color-spaces}

RGB color space is the most simple and familiar color system we've
ever known. Due to the basic and well-known properties, it's used
for most of color applications and follows the additive color law.
The primary components of RGB space are red, green, and blue
colors. By the additive color law, the secondary components are
cyan, magenta, and yellow which is obtained from the mixture of
the primary colors. The major problem that RGB model suffers from
is a strong degree of correlation among the three components. The
three values change dependently and are highly sensitive to the
variation of lightness~\cite{Che00}. In contrast to the RGB space,
CMY color space is mainly used for color reproduction industries,
for example, color printing works. The major principle in CMY
color system is the subtractive color law, which is the opposite
of the additive law of RGB space. The major primaries are cyan,
magenta, and yellow color.

\subsubsection*{2.1.2\quad Munsell Color Space}
\label{sec:munsell-color-space}

$Munsell$ $color$ $space$ is the most intuitive and useful to the
artists and designer~\cite{Lon94}. Every color sensation unites
three qualities, defined in the Munsell system as $hue$, $value$,
and $chroma$. These three components are independent of one
another because each can be varied without changing the others.

$Hue$ is the component which distinguishes one color family from
another, for example, red family from yellow family, or green
family from blue family. There are five major color families and
five minor hue families in the halfway between each major hues.
All hues are arranged in a circle. This circle of hues is often
called as a Munsell color wheel. $Value$ is a component by which a
light color is distinguished from a dark one, namely, it is a
metric for a lightness of a given color. The whole colors in the
value scale have no hue which is denoted by an $N$ (neutral). The
end points of value scale are true black (N 0/) and true white (N
1/). $Chroma$ is the component which expresses the strength of a
color. By the chroma level, we can distinguish the vividness
between two colors. A vivid color has strong chroma.

\subsubsection*{2.1.3\quad CIEXYZ Color Space}
\label{sec:ciexyz-color-space}

In 1931 CIE (International Commission on Illumination) established
the colorimetry science formally. From the time CIE has comprised
the essential standards and procedures of measurement that are
necessary to make colorimetry a useful tool in science and
technology. The CIE focused on the color gamut problem. The gamut
represents the entire range of color which can be achieved in a
specific device or medium. So we can easily plot the whole color
gamut that can be obtained from all color mediums. However the
problem CIE had noticed was that the RGB color space could not
reproduce all spectral light, which means it cannot cover the
whole range of color gamut. The CIE thought this happened in the
process of mixing red, blue, and green primaries, which caused a
negative value effect. So they transformed the RGB color space
into a different set of positive color stimuli values, called
CIEXYZ color space.

The transformed $XYZ$ color values are not exactly same to the
original red, green, and blue, but are approximately so. The $XYZ$
tristimulus values are just used for defining a color. But
unfortunately, we can not use this as the means for calculating
color difference formula and establishing human perception model
because CIEXYZ color is not uniform space. We can find that colors
of equal amounts of difference appear farther apart in green part
of the diagram rather than they do in the red or violet part. To
resolve the problem of non-uniform color scaling, CIE defined two
different uniform diagrams : CIELUV color space and CIELAB color
space.

\subsubsection*{2.1.4\quad CIELAB (1976) Color Space}
\label{sec:cielab-color-space}

A CIELAB color space is the second uniform space adopted by CIE in
1976 which showed a better uniform-scaling. The CIELAB space was
based on the opponent color system, which mentioned the
discoveries in the mid-1960s that somewhere between the optical
nerve and the brain, retinal color stimuli are translated into
distinctions between light and dark, red and green, and blue and
yellow. That means a color can't be both red and green, or both
blue and yellow, because these colors oppose each
other.~\cite{Ado00}. These opponent colors are translated to
CIELAB $L^*a^*b^*$ values. The $L^*$ value is the same as one of
CIE $L^*u^*v^*$ color space. Based on the opponent theory, two
axes in $a^*b^*$ space represent each opponent colors. On the
$a^*$ axis, positive $a^*$-value indicates amounts of red while
negative  $a^*$-value stands for the strength of green. In similar
way, On the $b^*$ axis, positive $b^*$-value indicates amounts of
yellow while negative $b^*$-value represents the blueness. For
both axes, zero is neutral gray which
 in $L^*$ axis.

The mathematical formula that calculates three $L^*,a^*,b^*$
values is defined as equation~\ref{eq:lab-1}~\cite{Wys00}.

\begin{equation}\label{eq:lab-1}
\begin{array}{l}
L^* = 116(Y/Y_{n})^{1/3}-16 \\
a^* = 500[(X/X_{n})^{1/3} - (Y/Y_{n})^{1/3}] \\
b^* = 200[(Y/Y_{n})^{1/3} - (Z/Z_{n})^{1/3}]
\end{array}
\end{equation}

with the constraint that $X/X_{n}$,$Y/Y_{n}$,$Z/Z_{n}>0.01$.

\subsubsection*{2.1.5\quad Fuzzy approach to color modeling}
\label{sec:fuzzy-approach-to-color-modeling}

Vertan, one of the pioneers in fuzzy color modeling, focused on
image processing problem~\cite{Ver01}. He is interested in the
imprecision and vagueness problem in digital image representation.
He said that noise, quantization and sampling errors, the
tolerance of the human visual system are the reasons of that
problem. He associated to each color $c$, that usually is a point
in the three-dimensional color gamut $C$, $\mu:C \rightarrow
[0,1]$ that measures the membership degree of any color $c\prime$
from $C$ within the class color $c$. Thus, $\mu_c(c\prime)$ is a
scalar within $[0,1]$ that expresses how similar is the color
$c\prime$ with respect to the color $c$.

The fuzzy membership computation model proposed by Vertan is

\begin{equation}
\mu_{c}(c\prime) =
    \left \{
    \begin{array}{lr}
        1.0 ~~~~~~~~~~~~~~~~~~~~~~~~~~~~~~\mbox{if}~~ d(c,c\prime) \leq JND \\
        max(0, 1.0-\frac{d(c,c\prime)}{\sigma JND}) ~~~~~~~ \mbox{if}~~ d(c,c\prime) > JND \\
    \end{array}
    \right.
\end{equation}

In the above equation, JND is the just noticeable color difference
for the CIELAB color space, and $\sigma$ is a tuning parameter to
modify the inter-color confusion.

\subsection*{2.2\quad Color Data Clustering}
\label{sec:color-data-clustering}

\subsubsection*{2.2.1\quad Clustering Task}
\label{sec:clustering-task}

Data clustering or cluster analysis organizes a collection of data
patterns which is usually represented as a vector of measurements
or a point in a multidimensional space into clusters based on a
kind of similarity measure\cite{Jai99}. In contrast to
discriminant analysis (or supervised classification), the
clustering task (often called an unsupervised classification)
classifies a given collection of unlabeled data into homogeneous
groups without any labeled patterns. All useful information is
obtained from the data itself. Clustering is a useful methodology
in various computer science field including decision-making,
machine-learning, data mining, document retrieval, image
segmentation, and pattern classification. Especially, it is very
meaningful in that there is little prior information available
about the data. Most of clustering research to deal with color
data, especially in color image segmentation, have handled the
color as a general crisp element without the consideration of
uncertainty and vagueness. Thus, popular clustering algorithms
with a conventional color, e.g. RGB color element, represented in
crisp color space have been just used.

The most intuitive and frequently used clustering algorithm is
partitional algorithm (often called K-Means algorithm). A
partitional clustering algorithm obtains a single partition of the
data in each iteration. The time and space complexities of the
partitional algorithms are typically lower than those of the
hierarchical-based algorithms (single-linkage and complete-linkage
algorithm) where two clusters are merged to form a larger cluster
based on minimum distance criteria. Thus, the partitional
algorithm has advantages in applications involving large data
sets, e.g. image segmentation, for which the construction of a
hierarchical linkage is computationally prohibitive~\cite{Jai99}.

The evaluation function of K-Means algorithm uses the squared
error criterion, which tends to work well with isolated and
compact clusters. The squared error for a clustering of a pattern
set $X$ (containing $c$ clusters) is
\begin{equation}
e^{2}(X) = \sum_{i=1}^{c}\sum_{j=1}^{n}(\|x_{j}^{i}-v_{i}\|)^2
\end{equation}

where $n$ is the number of data $x \in X$, and $x_{j}^{i}$ is the
$j^{th}$ pattern belonging to the $i^{th}$ cluster and $v_i$ is
the centroid of the $i^{th}$ cluster. The minimization of the
evaluation function generates a well-partitioned data collection.
However this is sensitive to the selection of the initial
partition and may converge to a local minimum of the criterion
function value if the initial partition is not properly chosen.

\subsubsection*{2.2.2\quad Fuzzy Clustering Algorithm}
\label{sec:fuzzy-clustering-algorithm}

The fuzzy clustering algorithm (often called FCM) generates a
fuzzy partition providing a measure of membership degree $\mu_{i}$
of each data $x_{i}$ to a given cluster $c_{i}$. The formal
description on Bezdek's fuzzy clustering algorithm is as follows.
For given data $x_1,...,x_n \in X$ where each $x_j$ is a
$p$-dimensional real-valued vector for all $k \in {1,...,n}$. The
goal is to find a partition or cluster of $c$ fuzzy sets $F_i,
i=1,...,c$. The objective is to assign memberships in each of the
fuzzy sets so that the data are strongly associated within each
cluster but only weakly associated between different clusters.
Each cluster is defined by its centroid, $v_i$, which can be
calculated as a function of the fuzzy memberships of all the
available data for the cluster.

The evaluation function ($J$) of fuzzy clustering looks like

\begin{equation}\label{eq:fcm-Jfunc}
J_{m}(F) =
\sum_{i=1}^{c}\sum_{j=1}^{n}[F_{i}(x_j)]^{m}\|x_j-v_i\|^2
\end{equation}

where $\| \cdot \|$ is an inner product norm in $\Re^p$ and
$\|x_j-v_j\|^2$ is a distance between $x_k$ and $v_i$. The
objective is to minimize $J_{m}(F)$ for some set of clusters $F$
and chosen value of $m$. The parameter $m > 1$ controls the
influence of the fuzzy membership of each datum. The choice of $m$
is subjective, however, many applications are performed using
$m=2$. That is, the goal is to iteratively improve a sequence of
sets of fuzzy clusters $F(1),F(2),...$ until a set $F(t)$ is found
such that no further improvement in $J_{m}(F)$ is possible. In
general, the design of membership functions is the most important
problem~\cite{Mic00}.

The advantages of fuzzy clustering algorithms can be summarized
like this~\cite{Sha00}. First, fuzzy clustering approach is less
prone to local minima than crisp clustering algorithms since they
make soft decisions in each iteration through the use of
membership functions. Second, the use of the fuzzy sets allows us
to manage uncertainty on measures, lack of information, ..., all
characteristics which bring ambiguity notions. Finally, in various
fields, such as pattern recognition, data analysis and image
processing, FCM is very robust and obtains good results in many
clustering problems. The algorithm provides an iterative
clustering of the search space and does not require any initial
knowledge about the structure in the data set.

\subsection*{2.3\quad Limitations of Previous Research}
\label{sec:limitations-of-previous-research}

\subsubsection*{2.3.1\quad Lack of Color Uncertainty Representation}
\label{sec:lack-of-color-uncertainty-representation}

One of the major factors affecting color description is that color
depends on neighboring color stimuli in the observer's visual
field. Cognitive scientists explain this results from the
interactions between cone receptors in the retina. These phenomena
are called "simultaneous contrast" or "chromatic induction
effect"~\cite{Ber00}. This color description can be understood by
computing perceived color difference among given colors. We could
easily guess that the mathematical color model which determines
the color difference perceivable by different for an observer is
mainly dependent on the observer. The observer may judge the color
difference relatively and subjectively. This is the reason none of
the many color difference formula that have been proposed in the
literature over the past several decades is considered as a
sufficiently adequate solution of the problem. As we can see, we
cannot handle this color problem with the conventional color
models. Most of color models including RGB and CIELAB persues the
crisp color representation and color difference formula. The
simple use of the color representation does not explain the
similar perception and visual confusion of certain colors.

Even though Vertan recently proposed the fuzzy approach for color
modeling, his model has also several drawbacks. One of them is
there's no notion on simultaneous contrast in his model. As
already mentioned, color difference description is dependent on
surrounding neighboring colors relatively. He didn't consider
neighboring effect in the membership computation. The color
membership should be calculated relatively according to the
surrounding neighbor colors.

\subsubsection*{2.3.2\quad Lack of Clustering Algorithm for Fuzzy Color Data}
\label{sec:lack0of-clustering-algorithm-for-fuzzy-color-data}

Most clustering algorithms are designed for treating crisp color
data. The practical techniques are K-Means clustering or FCM
clustering for RGB color representation. With this approach,
however, we cannot handle the color uncertainty problem. Yang and
Liu recently proposed a class of fuzzy c-numbers clustering
procedures for fuzzy data where high-dimensional fuzzy vector data
are handled~\cite{Yan99}. But the fuzzy data with conical fuzzy
vector is somewhat theoretical and is not appropriate for color
representation. Thus we need to develop an fuzzy color model that
can explain the color uncertainty and relative color difference
formula. With the proposed color model, we developed an effective
fuzzy clustering algorithm for color data.


\vspace{-1mm}\section*{III. Proposed Fuzzy Color Model}
\label{sec:proposed-fuzzy-color-model}

\subsection*{3.1\quad Research Approach}
\label{sec:research-approach}

In order to solve color clustering problem, we should take the
following three factors into account. The first one is a pattern
representation including the number, type, and scale of the
features. The second one is the definition of a inter-pattern
proximity measure. The most famous measure is Euclidean distance
measure. Finally we should develop an efficient clustering
algorithm. In this paper, we proposed and used a new fuzzy color
model as a color pattern representation method, and defined two
distance metrics to solve the inter-color pattern proximity.
Finally by developing fuzzy clustering algorithm with fuzzy color,
we could obtain an acceptable solution.

A new fuzzy color model should be created in order to satisfy the
following four conditions.

\begin{itemize}
\item[~~~~(1)] proposed model must provide human perception-based color distance
metric
\item[~~~~(2)] proposed model must be uniform color space
\item[~~~~(3)] proposed model must resolve the vague color boundaries
\item[~~~~(4)] proposed model must compute the color membership degree
in a relative way
\end{itemize}

In order to satisfy the conditions 1 and 2, we take a CIELAB color
space as a basic color coordinate. The CIELAB color space was
originally developed to provide the color difference formula like
human perception distance based on Munsell color description, and
more, it is uniform space and covers the whole color gamut. To
fulfill the conditions 3 and 4, we created a color model based on
fuzzy set theory. Thus color can be described as a membership
degree to a specific color, and the notion of color membership
resolves the vague boundaries between colors and relative color
acceptability.

\subsection*{3.2\quad Description of Fuzzy Color Model}
\label{sec:description-of-fuzzy-color-model}

\subsubsection*{3.2.1\quad Three-dimensional Fuzzy Color Ball}
\label{sec:three-dimensional-fuzzy-color-ball}

Fuzzy color is described with three dimensional ball-type
representation. To describe color concept in three dimensional
color space, sphere or ball-shaped model is preferred. In this
paper, a terminology $ball$ and $sphere$ are used interchangeably.
The formal definition of fuzzy color ball is described in section
3.2.3. When we look a color or colored object, for example red
color, some pairs of red colors are difficult to distinguish, and
beyond a certain boundary, we can easily distinguish color pairs.
Color has three dimensional volume representation. The radius of
color ball is the JND of each color. For a pair of two colors
within JND value, we assumed it is not easy to distinguish and we
consider those two colors are equal. And for colors which don't
belong to JND volume, those two colors can be distinguished.

\begin{figure}[t]
\begin{center}
\includegraphics[width=9cm]{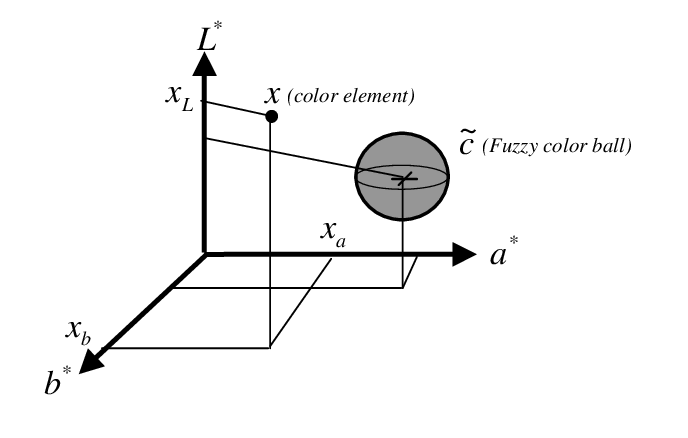}
\end{center}
\caption{A Fuzzy color ball on CIELAB color space}
\label{fig:fuzzy-ball-on-cielab}
\end{figure}

To describe the fuzzy color ball, two numerical values should be
specified: center and JND value. Center value is the center point
of fuzzy color ball which is calculated from CIELAB color space
and Munsell color wheel. And JND means a just noticeable distance
which is the distance from the center to a boundary of a given
color. Figure~\ref{fig:fuzzy-ball-on-cielab} shows a fuzzy color
ball on CIELAB color space. Here we must distinguish two objects:
color element and fuzzy color. Color element denoted by $x$ is a
point on CIELAB color coordinates, which is represented by $(x_L,
x_a, x_b)$ tuple value. Fuzzy color denoted by $\tilde{c}_i \in
\tilde{C}$ is a ball or sphere with a three dimensional volume
representation on CIELAB space. By taking CIELAB as a basic color
coordinates, it is no doubt that the proposed fuzzy color model
has an uniform color scaling.

\begin{figure}[b]
\begin{center}
\includegraphics[width=7cm]{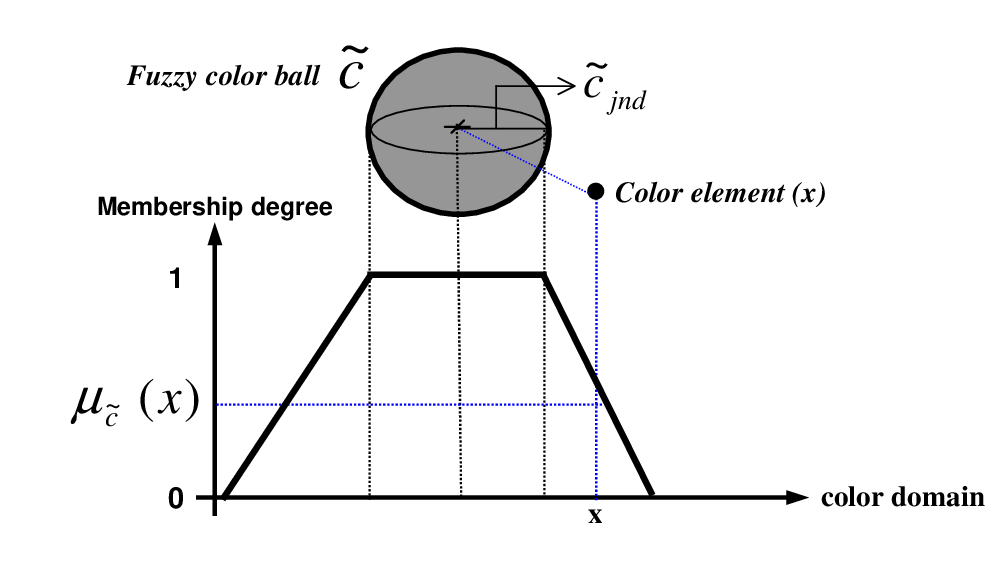}
\end{center}
\caption{Fuzzy color ball and its membership shape}
\label{fig:fuzzy-ball-and-membership-shape}
\end{figure}

For a given color element $x$, the membership
$\mu_{\tilde{c}_i}(x)$ of $x$ to fuzzy color $\tilde{c}_i$ is
obtained with the distance computation.
Figure~\ref{fig:fuzzy-ball-and-membership-shape} depicts the fuzzy
color ball and its membership function shape. In this figure,
there are fuzzy color $\tilde{c}_i$ and color element $x$. The
fuzzy color has its own center value $center_{i}$ and JND value
$jnd_{i}$ that build three dimensional ball-shaped representation.
If a color $x$ is within JND distance it strongly belongs to that
color and has membership degree 1.0. If color $x$ is out of the
range JND, the membership degree is computed relatively by
comparing with neighbor colors. The left and right shape of fuzzy
membership function is determined based on the distance result
between fuzzy colors.

\begin{figure}[b]
\begin{center}
\includegraphics[width=9cm]{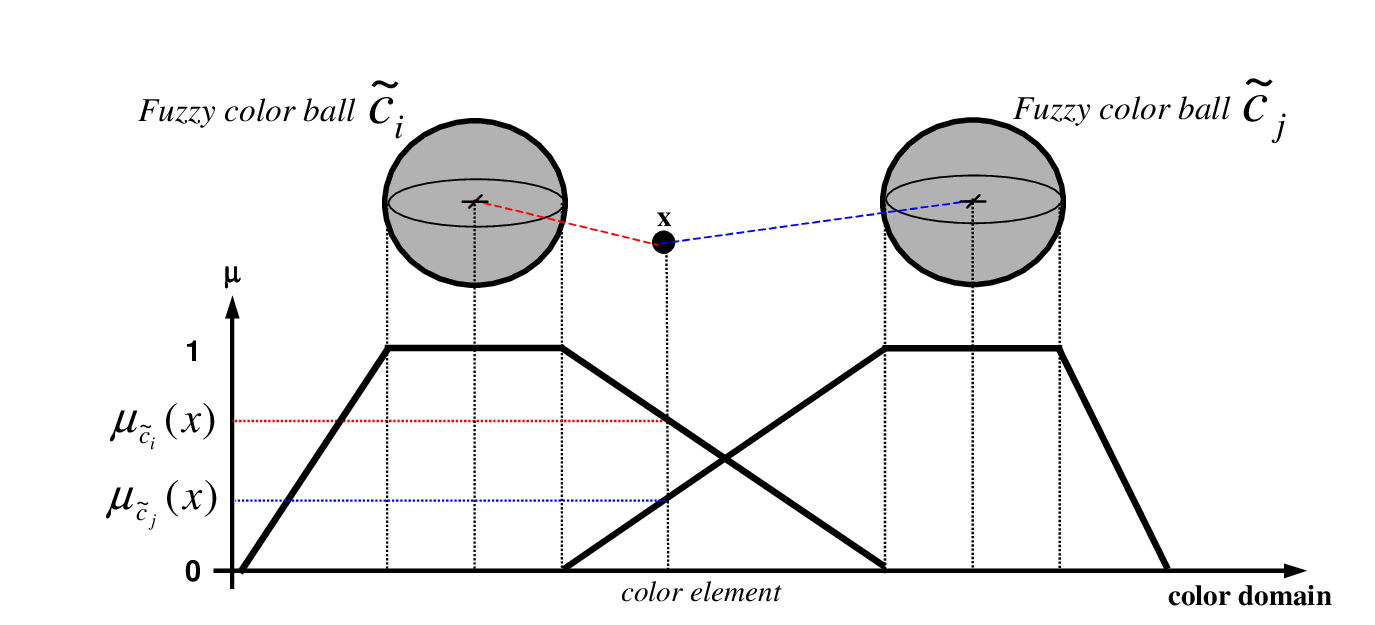}
\end{center}
\caption{Membership computation among fuzzy colors}
\label{fig:membership-among-fuzzy-colors}
\end{figure}

Figure~\ref{fig:membership-among-fuzzy-colors} depicts the
membership computation situation between two colors. To compute a
membership degree to a specific fuzzy color, we compute all
distances between fuzzy colors. If a color $x$ strongly belongs to
a given color $\tilde{c}_i$, it is classified to that color with a
membership degree $1.0$. In another case, if the color $x$ is
within other color $\tilde{c}_j$, then it means color $x$ has no
relation to the fuzzy color $\tilde{c}_i$, thus the membership
degree to $\tilde{c}_i$ is $0.0$. Except the above two cases, the
color $x$ is located in the middle of fuzzy colors. We compute the
relative distance between fuzzy colors and detemine the membership
value. And the important point is that sum of membership value to
the whole fuzzy colors must be equal to 1.0. With this constraint
we can easily extend this concept to fuzzy clustering algorithm.

\subsubsection*{3.2.2\quad Distance Measures Between Fuzzy Colors}
\label{sec:distance-measures-between-fuzzy-colors}

To effectively address the new definition of fuzzy color ball and
its membership computation method, we should define two
inter-color distances: distance between color elements and
distance between a fuzzy color and a color element.

\begin{figure}[b]
\begin{center}
\includegraphics[width=9cm]{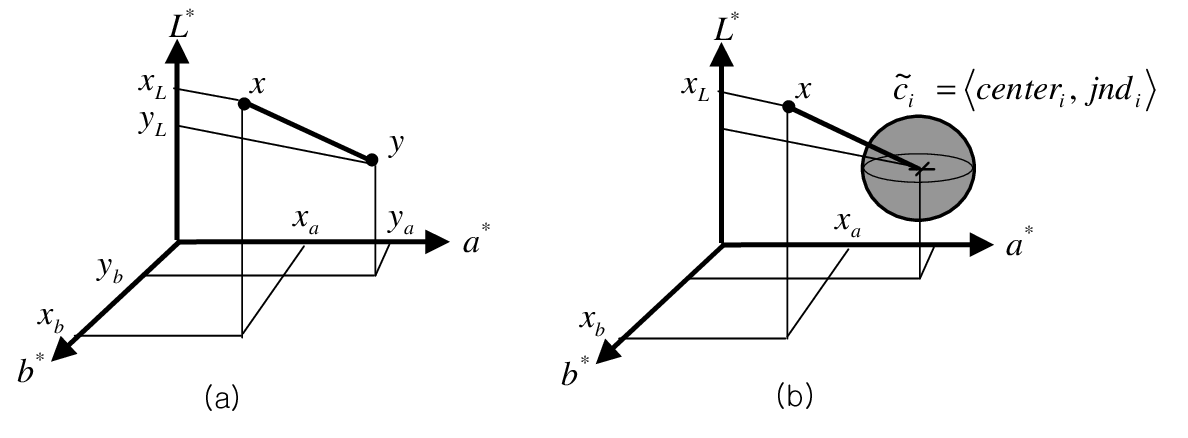}
\end{center}
\caption{Distance between color elements (a) and distance between
a fuzzy color and a color element (b) }
\label{fig:distance-figure}
\end{figure}

To successfully formulate the distance between fuzzy colors, a
basic color distance measure must be established between color
elements. \vspace{-9mm}$$$$
\begin{definition}\label{definition:distance-between-color-points}
Let $x$ and $y$ be two color elements on CIELAB color space, then
the distance between color element $x$ and color element $y$,
denoted by $\rho(x,y)$, is defined as
\begin{equation}
\rho(x,y) = \sqrt{(x_L-y_L)^2+(x_a-y_a)^2+(x_b-y_b)^2}
\end{equation}
where $x = (x_L,x_a,x_b)$ and $y = (y_L,y_a,y_b)$
\end{definition}\vspace{-7mm}
$$$$
The above definition is trivially obtained from $CIELAB$ color
difference formula. Figure~\ref{fig:distance-figure} (a) shows the
distance between color elements $x$ and $y$.

One of the major concerns is to compute the distance between a
fuzzy color and an arbitrary color element. With the distance
measure between color elements, we can define the distance between
a fuzzy color and a color element.
Figure~\ref{fig:distance-figure} (b) shows the distance between a
fuzzy color $\tilde{c_i}$ and a color element $x$.\vspace{-7mm}
$$$$
\begin{definition}\label{definition:distance-between-fuzzy-color}
Let $\tilde{c_i}$ and $x$ be a fuzzy color and a color element on
CIELAB color space respectively. Then the distance between fuzzy
color $\tilde{c_i}$ and color element $x$, denoted by
$\delta(\tilde{c_i}, x)$ is defined as
\begin{equation}
\begin{array}{l}
\delta(\tilde{c_i}, x) =
\| \rho(center_{i}, x) \| - jnd_{i} \\[0.1cm]
 = \sqrt{(center^{i}_{L}-x_L)^2+(center^{i}_{a}-x_a)^2+(center^{i}_{b}-x_b)^2}
\\[0.1cm]
\qquad\qquad - jnd_{i}
\end{array}
\end{equation}
where $center_i$ and $jnd_i$ are the center and JND value of fuzzy
color $\tilde{c_i}=\langle
(center^{i}_{L},center^{i}_{a},center^{i}_{b}),jnd_i \rangle$, and
$x = (x_L,x_a,x_b)$.
\end{definition}\vspace{-10mm}
$$$$
As can be seen in the above definition, the distance measure
considers not only the center point but also JND value. With the
$JND$ value, we can reflect the specific color's characteristics.
Some colors have larger JND values, and others have smaller ones
on color space. For example, greenish colors have bigger volume
than redish colors. Thus the adoption of JND values provides a
more clear distance computation result. Look at the following
figure~\ref{fig:different-jnd}. In the figure, if you compute the
color distances without the notion of JND value, the three
distance results from $x$ to fuzzy color ${\tilde{c}}_{red}$,
${\tilde{c}}_{blue}$, and ${\tilde{c}}_{green}$ would be equal
regardless to left and right cases of the figure. However, the
computation with JND value would generate different distance
results for left and right side of the figure. In the
figure~\ref{fig:different-jnd} (a), the two distances
$\delta({\tilde{c}}_{red},x)$ and $\delta({\tilde{c}}_{blue},x)$
are equal. In the figure~\ref{fig:different-jnd} (b), the distance
between ${\delta(\tilde{c}}_{green},x)$ is shorter than that of
$\delta({\tilde{c}}_{red},x)$ due to the JND value of fuzzy color
${\tilde{c}}_{green}$ occupies a wider area on color space. Thus
we can say that color element $x$ is closer to fuzzy color
${\tilde{c}}_{green}$ rather than ${\tilde{c}}_{red}$.

\begin{figure}[b]
\begin{center}
\includegraphics[width=8cm]{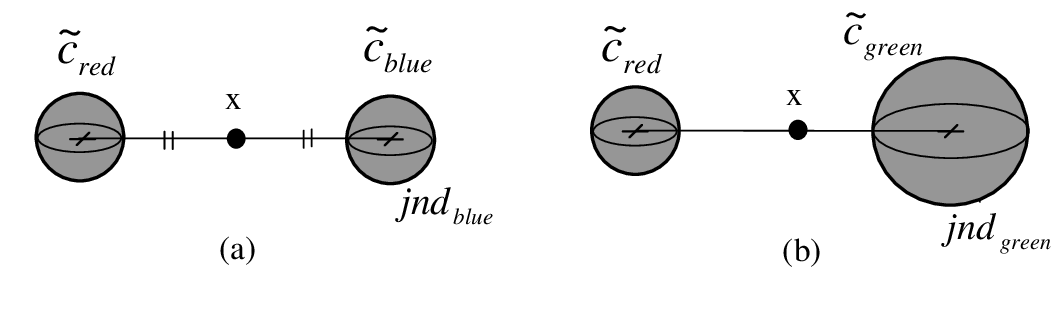}
\end{center}
\caption{Distance computation of fuzzy colors with different JND
value} \label{fig:different-jnd}
\end{figure}

\subsubsection*{3.2.3\quad Definition of Fuzzy Color Ball}
\label{sec:definition-of-fuzzy-color-ball}

With the distance computation measures defined in the above
section, we can establish a formal fuzzy color model.\vspace{-3mm}
$$$$
\begin{definition}\label{definition:revised-fuzzy-color}
Let fuzzy color $\tilde{c}_i$ in a universe of discourse
$\tilde{C}$ be as a three dimensional fuzzy ball set
$\tilde{c_i}=\langle center_i, jnd_i \rangle$ $\in \tilde{C}$ with
a membership function such that
\begin{equation}
\mu_{\tilde{c_i}}(x)= \left \{
            \begin{array}{lr}
                1.0 ~~~~~~~~~~~~~~~~~~~~\mbox{if}~~ \delta(\tilde{c_i}, x) \leq jnd_i \\
                0.0 ~~~~~~~~~~~~~~~~~~~~\mbox{if}~~ \delta(\tilde{c_j}, x) \leq jnd_j \\[0.1cm]
                                     \qquad  \qquad \qquad  \qquad \qquad (i \neq j, \tilde{c_j} \in \tilde{C})\\
                (\sum_{j=1}^{|\tilde{C}|}\frac{\delta(\tilde{c_i}, x)}{\delta(\tilde{c_j}, x)})^{-1}
                ~~~otherwise
            \end{array}
            \right.
\end{equation}
where $\delta$-function means the distance between a fuzzy color
and a color element.
\end{definition}\vspace{-10mm}
$$$$
To compute a membership degree to a specific fuzzy color, we
compute all distances between fuzzy colors. If a color $x$
strongly belongs to a given fuzzy color $\tilde{c_i}$, it is
classified to that color with a membership degree 1.0. In another
case, if the color $x$ is within other fuzzy color $\tilde{c_j}$,
then it means color $x$ has no relation to the fuzzy color
$\tilde{c_i}$, so the membership degree is 0.0. Except the above
two cases, the color $x$ is located in the middle of fuzzy colors.
We compute the relative distance between fuzzy colors and
determine the membership value. And the important point is that
the sum of membership values to the whole fuzzy color families
must be equal to 1.0. With this constraint we can easily extend
this concept to fuzzy cluster analysis which is discussed in later
section. The properties of the proposed fuzzy color model can be
summarized as follows:

\begin{itemize}
\item[~~~~(1)] If a color element $x$ absolutely belongs to a fuzzy color
$\tilde{c_i}$, then $\mu_{\tilde{c_i}}(x)=1$.
\item[~~~~(2)] If a color element $x$ absolutely belongs to a fuzzy color
$\tilde{c_j},i \neq j$, then $\mu_{\tilde{c_i}}(x)=0$.
\item[~~~~(3)] If a color element $x$ is partially related to the surrounding neighbor
fuzzy colors, then $0<\mu_{\tilde{c_i}}(x)<1$.
\item[~~~~(4)] If a fuzzy color $\tilde{c_i} \in \tilde{C}$ is the closest
fuzzy color to color element $x$, then $\mu_{\tilde{c_i}}(x) \geq
\mu_{\tilde{c_j}}(x) (\forall \tilde{c}_j \in \tilde{C})$.
\item[~~~~(5)] The sum of all membership values of color element $x$ to whole
fuzzy colors is equal to 1.0.
($\sum_{i=1}^{|\tilde{C}|}\mu_{\tilde{c_i}}(x)=1.0$)
\end{itemize}

\subsubsection*{3.2.4\quad Numerical Example}
\label{sec:numerical-example}

In this paragraph, we compute the fuzzy color distance with
numerical examples. In the following examples, there are three
fuzzy colors including ${\tilde{c}}_{red}$, ${\tilde{c}}_{green}$,
and ${\tilde{c}}_{blue}$, and one color element $x$. We calculate
the distance between fuzzy color ${\tilde{c}}_{red}$ and a given
color element $x$. Let's compute the membership degree of $x$ to
fuzzy color ${\tilde{c}}_{red}$.

\begin{figure}[b]
\begin{center}
\includegraphics[width=9cm]{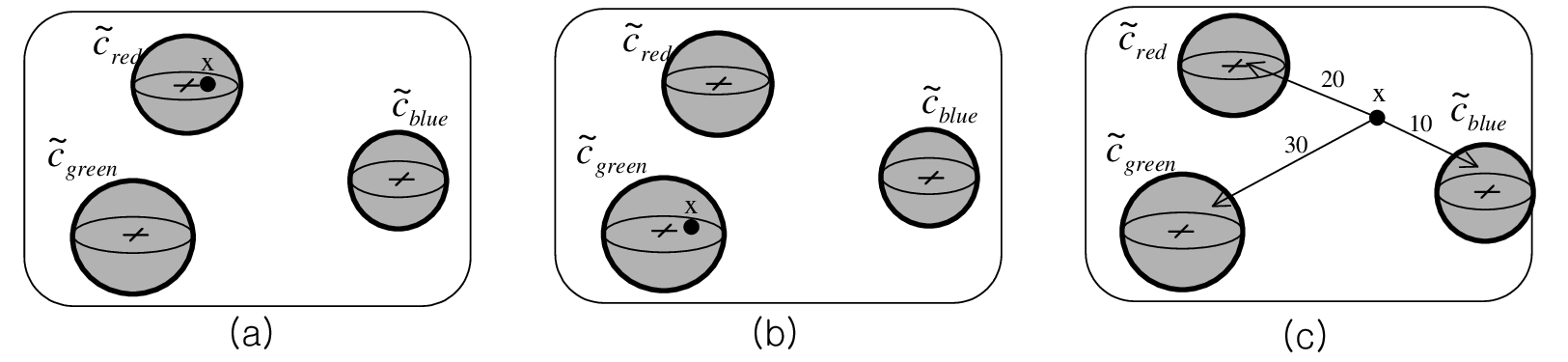}
\end{center}
\caption{Numerical example of membership computation}
\label{fig:example}
\end{figure}

In figure~\ref{fig:example} (a), color element $x$ belongs to the
color ${\tilde{c}}_{red}$ perfectly. In this case the membership
degree $\mu_{{\tilde{c}}_{red}}(x)$ is 1.0
($\mu_{{\tilde{c}}_{green}}(x)=0.0$ and
$\mu_{{\tilde{c}}_{blue}}(x)=0.0$). In figure~\ref{fig:example}
(b), color element $x$ belongs to the color family
${\tilde{c}}_{green}$ perfectly. In this case color element $x$
has no relations to fuzzy color ${\tilde{c}}_{red}$. Thus the
membership degree $\mu_{{\tilde{c}}_{red}}(x)$ is 0.0
($\mu_{{\tilde{c}}_{green}}(x)=1.0$ and
$\mu_{{\tilde{c}}_{blue}}(x)=0.0$). Look at the
figure~\ref{fig:example} (c), where color element $x$ doesn't
belong to any fuzzy colors at all, and the distances to each fuzzy
color are given like this: 20 to ${\tilde{c}}_{red}$, 30 to
${\tilde{c}}_{green}$, and 10 to ${\tilde{c}}_{blue}$. In this
case, we should first compute the relative strengths between fuzzy
colors, and based on the strength we compute the relative color
membership values. The result memberships are calculated as:

$$
\mu_{{\tilde{c}}_{red}}(x) =(
\frac{\delta({\tilde{c}}_{red},x)}{\delta({\tilde{c}}_{red}, x)} +
\frac{\delta({\tilde{c}}_{red},x)}{\delta({\tilde{c}}_{green},
x)}+ \frac{\delta({\tilde{c}}_{red},x)}{\delta({\tilde{c}}_{blue},
x)})^{-1}
$$
$$
\mu_{{\tilde{c}}_{green}}(x) =(
\frac{\delta({\tilde{c}}_{green},x)}{\delta({\tilde{c}}_{red}, x)}
+ \frac{\delta({\tilde{c}}_{green},x)}{\delta({\tilde{c}}_{green},
x)}+
\frac{\delta({\tilde{c}}_{green},x)}{\delta({\tilde{c}}_{blue},
x)})^{-1}
$$
$$
\mu_{{\tilde{c}}_{blue}}(x) =(
\frac{\delta({\tilde{c}}_{blue},x)}{\delta({\tilde{c}}_{red}, x)}
+ \frac{\delta({\tilde{c}}_{blue},x)}{\delta({\tilde{c}}_{green},
x)}+
\frac{\delta({\tilde{c}}_{blue},x)}{\delta({\tilde{c}}_{blue},
x)})^{-1}
$$

Thus the final membership degree of color $x$ to fuzzy colors are
like this:

$$
\mu_{{\tilde{c}}_{red}}(x) = (20/20 + 20/30 + 20/10)^{-1} = 0.27
$$
$$
\mu_{{\tilde{c}}_{green}}(x) = (30/20 + 30/30 + 30/10)^{-1} = 0.18
$$
$$
\mu_{{\tilde{c}}_{blue}}(x) = (10/20 + 10/30 + 10/10)^{-1} = 0.55
$$

With the result, we can say that color element $x$ is close to
fuzzy color ${\tilde{c}}_{red}$ with a membership 0.27. As
mentioned earlier, the membership constraint (the sum of
membership to all fuzzy color families is 1.0) in examples is
successfully satisfied.


\vspace{-1mm}\section*{IV. Fuzzy Clustering Algorithm based on
Fuzzy Color Model}
\label{sec:fuzzy-clustering-based-on-fuzzy-color-model}

\subsection*{4.1\quad Problem Formulation of Color Clustering}
\label{sec:problem-formulation-of-color-clustering}

The goal of our research is to cluster the color space represented
in CIELAB color system. To accomplish this, fuzzy clustering
technique is selected as a major algorithm approach because the
major feature 'color' of the data pattern has uncertain and vague
properties. Thus fuzzy approach has an advantage over the hard
clustering algorithms to handle these problems.

The following terms and notations are used to describe the problem
formulation. The given color element data are $x_1,...,x_n \in X$
where each $x_j$ ($j \in {1,...,n}$) is a color element
represented as three-dimensional $L^*a^*b^*$ value on CIELAB color
space. The algorithm objective is to cluster a collection of given
color elements into $c$ homogeneous groups represented as fuzzy
sets ($\tilde{F_i}, i=1,...,c$) with similar color
characteristics. The fuzzy cluster set can be written like an
equation~\ref{eq:fuzzy-cluster-set}.
\begin{equation}
\label{eq:fuzzy-cluster-set} \begin{array}{l}\tilde{F_i}=\{(x_1,
\mu_{F_i}(x_1), (x_2, \mu_{F_i}(x_2),..., (x_n,
\mu_{F_i}(x_n)\}\\[0.1cm]
\,\quad =\sum_{j=1}^{n}\mu_{F_i}(x_j)/x_j
\end{array}
\end{equation}
 The collection of fuzzy cluster set is denoted by
$\tilde{F}=\{\tilde{F_1},\tilde{F_2},..,\tilde{F_c}\}$. Each fuzzy
cluster is defined by its centroid, denoted by $\tilde{c}_i$,
which is represented by the proposed fuzzy color model. The
pattern matrix $M$ which is handled by clustering algorithm is
represented as an $c \times n$ pattern matrix. An element $p_{ij}$
in matrix $M$ means that a color element $x_j$ is classified to a
fuzzy cluster $\tilde{F}_i$.

The equation~\ref{eq:eval-fn} shows the evaluation function of the
proposed approach. The objective is to minimize the evaluation
function $J(\tilde{F})$ for a given pattern matrix.
\begin{equation}
\label{eq:eval-fn} J(\tilde{F}) =
\sum_{i=1}^{c}\sum_{j=1}^{n}[\mu_{\tilde{F}_i}(x_j)]^{2}\delta(\tilde{c_i},x_j)
\end{equation}
where $\mu_{\tilde{F}_i}(x_j)$ is the membership degree of color
element $x_j$ to a fuzzy cluster set $\tilde{F}_i$, and
$\delta(\tilde{c_i},x_j)$ is the distance between the color
element $x_j$ and the centroid $\tilde{c_i}$ of the fuzzy cluster
set $\tilde{F}_i$. The goal is to iteratively improve a sequence
of sets of fuzzy clusters $\tilde{F}(1),\tilde{F}(2),...$ until
$\tilde{F}(t)$ is found such that no further improvement in
$J(\tilde{F})$ is possible. In general, the design of membership
function and centroid prototype are the most important problems.
The detailed algorithm is discussed in
section~\ref{sec:skeleton-of-proposed-fuzzy-clustering-algorithm}.

\subsection*{4.2\quad Initialization of Cluster Analysis}
\label{sec:initialization-of-clustet-analysis}

Before we describe the detailed clustering algorithm, we discuss
how to obtain the initial partition of color cluster set. In
cluster analysis, initialization plays an important role.
According to the initial starting point, the clustering algorithm
might terminate at different clustering result. There is no
general agreement about a good initialization scheme. Two most
popular techniques are (1) using the first $c$ pattern data (2)
using $c$ elements randomly from pattern set. In this paper, we
propose a novel initial selection method based on the notion of
fuzzy color. It is simple and intuitive.

\begin{figure}[t]
\begin{center}
\includegraphics[width=9cm]{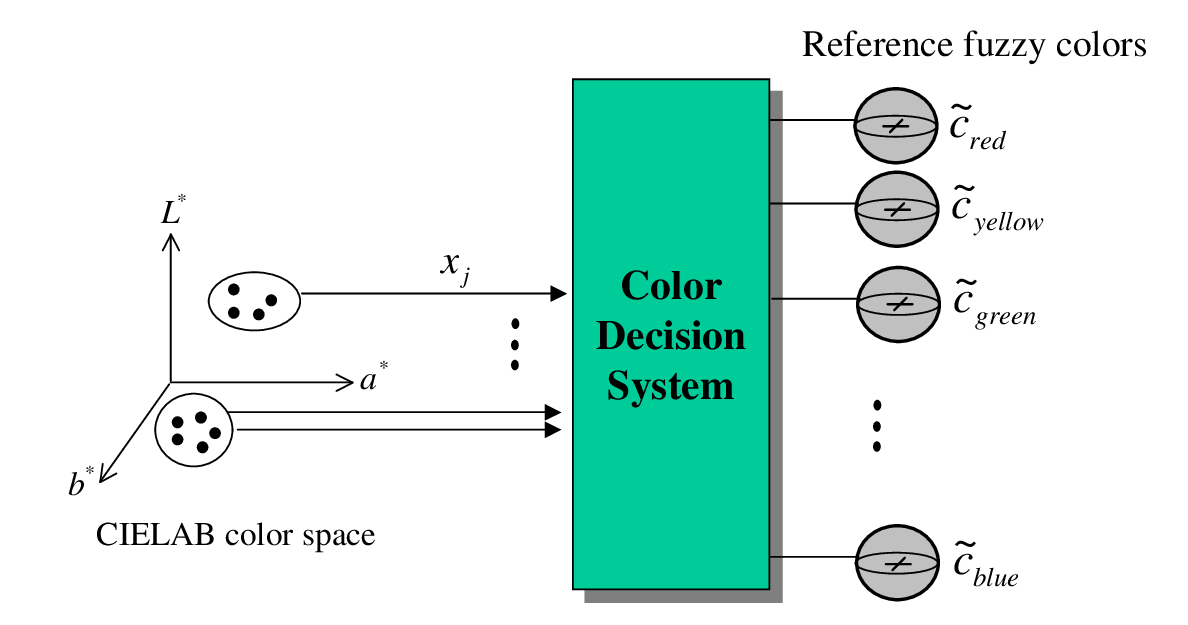}
\end{center}
\caption{Initial selection of fuzzy cluster centroid}
\label{fig:color-decision-system}
\end{figure}

The basic idea is as follows. For a color element $x_j \in X$, we
compute the matching degree to pre-defined thirteen reference
fuzzy colors which is denoted by $\tilde{c}_R$. Reference fuzzy
colors are selected from both the major color family in Munsell
color wheel (5R, 5YR, 5Y, ..., 5P, and 5RP) and the $L^*$-axis
components (white, black, gray) of CIELAB color space.
Figure~\ref{fig:color-decision-system} shows the overall initial
selection process. The input of a initial color decision system is
color elements in pattern space, and the output is a list of fuzzy
colors where the fuzzy colors are sorted by the maximum matching
score. The first $c$ fuzzy colors are chosen as the initial fuzzy
cluster centroid.

\vspace{-1mm} For a given color element $x_j$, the matching score
is computed by considering the membership degree
$\mu_{{\tilde{c}}_i}(x_j)$ to all reference fuzzy colors. Each
reference fuzzy color $\tilde{c}_i \in \tilde{C}_R$ has two
additional attributes denoted by $winner_{i}^{count}$ and
$winner_{i}^{element}$. The $winner_{i}^{count}$ means how many
color elements belong to the given reference fuzzy color
$\tilde{c}_i$.\vspace{-5mm}

\begin{equation}
\begin{array}{l}winner_i^{count} \longleftarrow winner_i^{count} + 1
\\[0.1cm]
\qquad \qquad ~~~~~if~~ \mu_{\tilde{c}_i}(x_j) \geq
\mu_{\tilde{c}_k}(x_j) (\forall \tilde{c}_k \in \tilde{C}_R)
\end{array}
\end{equation}

The $winner_{i}^{element}$ is used as a center element of a newly
created fuzzy color and is assigned as the color element that has
maximum membership degree to this reference fuzzy
color.\vspace{-5mm}

\begin{equation}
\begin{array}{l}
winner_i^{element} \longleftarrow x_j\\[0.1cm]
\qquad  ~~~~~~~~~~~~~~if~~ \mu_{\tilde{c}_i}(x_j) \geq
\mu_{\tilde{c}_i}(x_k) (\forall x_k \in X)
\end{array}
\end{equation}

After the classification decision process, we build a sorted list
of thirteen fuzzy colors according to their $winner_{i}^{count}$
in a descending order. From the sorted list we select the first
$c$ (number of clusters) fuzzy colors which have a larger matching
counter than other colors. We don't need to focus on the whole
thirteen colors in the list, just $c$ fuzzy colors are enough for
the candidates for cluster centroid.

\subsection*{4.3\quad Fuzzy Clustering using Fuzzy Color}
\label{sec:fuzzy-clustering-using-fuzzy-color}

\subsubsection*{4.3.1\quad Fuzzy Colored Centroid Representation}
\label{sec:fuzzy-colored-centroid-representation}

The previous fuzzy clustering algorithm including FCM was designed
only to deal with crisp data element. This means they can be
regarded as a fuzzy clustering algorithm for crisp data. They did
not consider any mechanisms about how to handle fuzzy data,
especially color characteristics, even though the fuzzy clustering
methodology is very appropriate to efficiently deal with the vague
boundaries between fuzzy color data. Cluster analysis can be
explained in the viewpoint of pattern matrix $M$. By updating the
pattern matrix iteratively in each sequence step, clustering
algorithm goes on the convergence condition. The typical matrix
$M$ has a $c \times n$ matrix form, where each cluster centroids
$c_i$ represent the row attribute of $M$, and each data element
$x_j$ stands for the column of matrix. In this situation all the
data elements and cluster centroids have crisp representation,
thus conventional approach has a limitation to account for the
uncertainty and vagueness in color clustering.

\begin{figure}[t]
\begin{center}
\includegraphics[width=9cm]{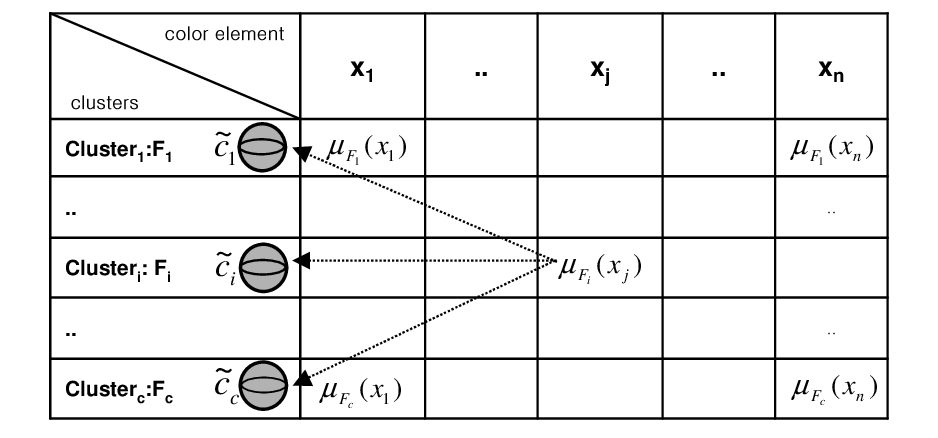}
\end{center}
\caption{Membership update in proposed pattern matrix}
\label{fig:proposed-clustering}
\end{figure}

We tried to represent the fuzziness and vagueness resolution
method into pattern matrix. Proposed approach is to replace the
typical point-prototype crisp centroid with a new cluster centroid
represented by the fuzzy color.
Figure~\ref{fig:proposed-clustering} depicts the skeleton of the
proposed pattern matrix. The column of matrix contains each color
elements $x_j$, and the row elements represent the centroid
$\tilde{c}_i$ of each fuzzy cluster $\tilde{F}_i$ which is shown
as three dimensional fuzzy color ball. Fuzzy color centroid has
its own center element and JND value that are a specific
properties of a given fuzzy cluster. This makes it possible to
cope with the clustering decision problem in the vague region
between neighbor fuzzy color clusters. With the initialization
step to obtain the initial fuzzy color centroids $\tilde{c}_i$,
the membership computation of each element
$\mu_{{\tilde{F}}_i}(x_j)$ in pattern matrix is carried out. The
membership degree of color element $x_j$ to the fuzzy cluster
$\tilde{F}_i$ is computed by the relative distance between
neighbor fuzzy color centroids. After this cluster initialization
process, the fuzzy color centroids and the membership degrees of
matrix elements are iteratively updated until no improvements are
found.

\subsubsection*{4.3.2\quad Skeleton of Proposed Algorithm}
\label{sec:skeleton-of-proposed-algorithm}

As mentioned in earlier section, the objective is to obtain the
optimal partition
$\tilde{F}=\{\tilde{F_1},\tilde{F_2},..,\tilde{F_c}\}$ for a given
color elements $x_1,...,x_n$ and number of clusters $c$ by
minimizing the evaluation function $J(\tilde{F})$. The algorithmic
step can be described as followings:

\begin{description}
\item[Step 1.] With a given pre-determined number of clusters ($c$) and the color data
elements ($x_1,...,x_n \in X$), run the proposed cluster
initialization method to obtain the candidates of fuzzy color
centroids.
\item[Step 2.] Create the initial $\tilde{c}$ fuzzy color centroids,
$\tilde{c_{1}}(t),...,\tilde{c_{c}}(t)$ where $t$ means the
iteration step (initially $t=0$) with the candidates obtained in
Step 1. The newly created fuzzy color centroid $\tilde{c}_i$ is
defined as follows:
\begin{equation}
\tilde{c_i}=\langle center_i, jnd_i\rangle
\end{equation}
\begin{equation}
center_i \leftarrow winner_{i}^{element}
\end{equation}
\begin{equation}
jnd_i \leftarrow jnd_k ~~s.t.~~ \{k|\forall \tilde{c}_k \in
\tilde{C}_R,~ min_k\delta(\tilde{c}_k, center_i)\}
\end{equation}
where $\delta$-function means the distance between a color element
and a fuzzy color, and $\tilde{C}_R$ represent the universe of
discourse of the predefined thirteen reference fuzzy colors.
\item[Step 3.] Update the membership degree of fuzzy cluster sets $\tilde{F}_{i}(t+1)$ by the following
procedure. For each color element $x_j$:

\begin{itemize}
\item[(a)] if $\delta(\tilde{c_i},x_j)<jnd_{\tilde{c_i}}$, then update the
membership of $x_j$ in $\tilde{F}_i$ at $t+1$ iteration by
\begin{equation}
\mu_{\tilde{F}_i}(x_j)(t+1) = 1.0
\end{equation}
\item[(b)] if $\delta(\tilde{c_k},x_j)<jnd_{\tilde{c_k}}$, for $k \neq i,\tilde{c}_k\in\tilde{C}$
then update the membership of $x_j$ in $\tilde{F}_i$ at $t+1$
iteration by
\begin{equation}
\mu_{\tilde{F}_i}(x_j)(t+1) = 0.0
\end{equation}
\item[(c)] if condition (a) and (b) are not satisfied for all $\tilde{c}_k\in\tilde{C},k=1..c$,
then update the membership of $x_j$ in $\tilde{F}_i$ at $t+1$
iteration by
\begin{equation}
\mu_{\tilde{F}_i}(x_j)(t+1) =
(\sum_{k=1}^{c}\frac{\delta(\tilde{c}_i,x_j)}{\delta(\tilde{c}_k,x_j)})^{-1}
\end{equation}
where the column sum constraint
$\sum_{i=1}^{c}\mu_{\tilde{F}_i}(x_j)=1.0$ should be satisfied.
\end{itemize}
\item[Step 4.] Update the fuzzy color centroid $\tilde{c_i}$ of each fuzzy cluster. The $jnd_i$ of each fuzzy color centroid
$\tilde{c_i}$ is updated in a similar way of step 2. The
$center_i$ value is computed by the following procedure.
\begin{equation}
center_i(t+1) = \frac{\sum_{j=1}^{n}\mu_{\tilde{F}_i}(x_j)(t)
\cdot x_j} {\sum_{j=1}^{n}\mu_{\tilde{F}_i}(x_j)(t)}
\end{equation}
\item[Step 5.] If $|\tilde{F}_i(t+1)-\tilde{F}_i(t)| < \epsilon$ for all $\tilde{F}_i\in\tilde{F}$, where $\epsilon$ is a
small positive constant, then halt because it's believed that
algorithm has reached at convergence; otherwise, $t \leftarrow
t+1$ and go to step 3.
\end{description}


\vspace{-1mm}\section*{V. Concluding Remarks}
\label{sec:concluding-remarks}

In this paper we discussed the color clustering problem. To
successfully partition the color pattern data, we first modeled a
new fuzzy color model that can describe the vagueness underlying
natural colors. The fuzzy color model is based on the CIELAB color
space which gives the uniform color scaling. We defined the
concept of fuzzy color and a new distance measure between fuzzy
color and a color point. Each fuzzy color has a tuple of its
center and JND value. The JND value describes the different
properties of a given color. With the fuzzy color distance we
proposed a color membership computation method to a specific fuzzy
color and its desirable properties. In order to effectively deal
with color data in clustering, we adopted a fuzzy cluster
analysis. Because the fuzzy clustering makes a soft decision in
each iteration through the use of membership functions, it may be
the best technique in processing the fuzzy color data. We
developed a new fuzzy clustering algorithm with the proposed fuzzy
color model. The key idea was to exploit the fuzzy color
centroids. Each fuzzy color centroid can help to calculate the
membership degree of each color data.

Based on the current work, we would like to study the following
issues as a further work. We should take an analysis and a
mathematical proof on the proposed fuzzy color model, its
properties. And the color's asymmetric characteristics and the
reference color selection policy should be considered. As a
clustering issue, the most important keys include the automatic
determination of the number of clusters and centroid
initialization technique. In addition to this point, we should
check that the partitional clustering approach with
point-prototype representation is a really good solution.
Especially the time complexity is computationally prohibitive for
fuzzy clustering. And the possibility of transformation of color
point to fuzzy color representation is also one of our concern.
The hardening issue in fuzzy clustering is also very important
because it is closely related to the defuzzification of fuzzy
cluster sets.


\label{last}


\begin{thebibliography}{99}

\bibitem{Ado00}
Adobe Systems. (2000). {\em Technical Guides: Color Models.\/} In
the hompage of http://www.adobe.com/.

\bibitem{Ben96}
Bensaid, A.M., Hall, L.O., Bezdek, J.C, and Clarke, L.P. (1996).
Partially Supervised Clustering for Image Segmentation. {\em
Pattern Recognition,\/} Vol.29, No.5, pp.859--871.

\newpage\bibitem{Bez81}
Bezdek, J.C. (1981). {\em Pattern Recognition with Fuzzy Objective
Function Algorithm.\/} Plenum Press, New York.

\bibitem{Che00} Cheng, H.D. (2000). A hierarchical approach to
color image segmentation using homogeneity. {\em IEEE Transactions
on image processing,\/} Vol.9, No.12, pp.2071--2082.s

\bibitem{Gow92}
Gowda, K.C. and Diday, E. (1992). Symbolic clustering using a new
dissimilarity measure. {\em IEEE Transactions on Systems, Man and
Cybernetics,\/} Vol.22, pp.368--378.

\bibitem{Jai88} Jain, A.K. and Dubes, R.C. (1988). {\em
Algorithms for Clustering Data.\/} Prentice-Hall, NJ.

\bibitem{Jai99}
Jain, A.K., Murty., M.N., and Flynn, P.J. (1999). Data Clustering:
A Review {\em ACM Computing Surveys,\/} Vol.31, No. 3,
pp.264--323.

\bibitem{Kri93}
Krishnapuram, R. (1993). A possibilistic approach to clustering.
{\em IEEE Transactions on Fuzzy Systems,\/} Vol.1, No.2,
pp.98--110.

\bibitem{Kur91}
Kurita, T. (1991). An efficient agglomerative clustering using
heap. {\em Pattern Recognition,\/} Vol.24, No.3, pp.205--209.

\bibitem{Lon94}
Long, J. and Luke, J.T. (1994). {\em New Munsell Student Color
Set.\/} Fairchild Publications, New York.

\bibitem{Luc99}
Lucchese, L. and Mitra, S.K. (1999). Advances In Color Image
Segmentation. {\em 1999 IEEE Global Telecommunications
Conference,\/} pp. 2038--2044.

\bibitem{Luk96}
Luke, J.T. (1996). {\em The Munsell color system: a language for
color.\/} Fairchild Publications, New York.

\bibitem{Mic00}
Michalewicz, Z. and Fogel, B. (2000). {\em How to Solve It: Modern
Heuristics.\/} Springer-Verlag, Berlin.

\bibitem{Pal93}
Pal, N.R. and Pal, S.K (1993). A Review on Image Segmentation
Techniques. {\em Pattern Recognition,\/} Vol.26, No.9,
pp.1277--1294.

\bibitem{Par97}
Park, S.H, Yun, I.D, and Lee, S.U. (1998). Color Image
Segmentation Based on 3-D Clustering: Morphological Approach. {\em
Pattern Recognition,\/} Vol.31, No.8, pp.1061--1076.

\bibitem{Sha97}
Sharma, G. and Trussell, H.J. (1997). Digital Color Imaging. {\em
IEEE Transactions on Image Processing,\/} Vol.6, No.7,
pp.901--932.

\bibitem{Sha00}
Shahin, A., Menard, M., and Eboueya, M. (2000). Cooperation of
fuzzy segmentation operators for correction aliasing phenomenon in
3D color doppler imaging. {\em Artificial Intelligence,\/} Vol.19,
No.2, pp.121--154.

\bibitem{Won01}
Wong, C.C, Chen, C.C, and Su, M.C. (2001). A Novel Algorithm for
Data Clustering. {\em Pattern Recognition,\/} Vol.34, pp.425-442.

\bibitem{Wys00}
Wyszecki, G. and Stiles, W.S. (2000). {\em Color Science :
Concepts and Methods, Quantitative Data and Formulae.\/}
Wiley-Interscience Publication, New York.

\bibitem{Ber00}
Berns, R.S. (2000). {\em Principles of Color Technology.\/}
Wiley-Interscience Publication, New York.

\bibitem{Xie91}
Xie, X.L. and Beni, G. (1991). A Validity Measure for Fuzzy
Clustering. {\em IEEE Transactions on Pattern Analysis and Machine
Intelligence ,\/} Vol.13, No.8, pp.841--847.

\bibitem{Yan99}
Yang, M.S. and Liu, H.H (1999). Fuzzy Clustering Procedures for
Conical Fuzzy Vector Data. {\em Fuzzy Sets and Systems,\/}
Vol.106, pp.189--200.

\bibitem{Zel98}
Zelanski, P. and Fisher, M.P. (1998). {\em Color.\/} Calmann and
King Ltd.

\bibitem{Ver01}
Vertan, C, Stoica, A, and Fernandez-Malogine. (2001). Perceptual
Fuzzy Multiscale Color Edge Detection. {\em IEEE International
Conference on Fuzzy Systems}

\end{thebibliography}
\end{document}